\documentclass{article}

\usepackage{PRIMEarxiv}

\usepackage[utf8]{inputenc} 
\usepackage[T1]{fontenc}    
\usepackage{hyperref}       
\usepackage{url}            
\usepackage{booktabs}       
\usepackage{adjustbox}
\usepackage{amsfonts}       
\usepackage{nicefrac}       
\usepackage{microtype}      
\usepackage{lipsum}
\usepackage{fancyhdr}       
\usepackage{graphicx}       
\graphicspath{{media/}}     
\usepackage{xcolor}
\usepackage{soul}
\usepackage{tcolorbox}
\tcbuselibrary{listingsutf8}
\usepackage[edges]{forest}
\usepackage{comment}

\title{eLasmobranc Dataset: An Image Dataset for Elasmobranch Species Recognition and Biodiversity Monitoring
}

\author{
Ismael Beviá-Ballesteros*$^{1,2}$,
Mario Jerez-Tallón$^{1,2}$,
Nieves Aranda-Garrido$^{1,3}$,
Isabel Abel-Abellán$^{1,3}$,\\
\textbf{Irene Antón-Linares$^{1,3}$,
Jorge Azorín-López$^{1,2}$,
Marcelo Saval-Calvo$^{1,2}$,
Andres Fuster-Guilló$^{1,2}$,}\\
\textbf{Francisca Giménez-Casalduero$^{1,3}$} \\
\\
$^1$University of Alicante \\
$^2$Department of Computer Science and Technology (DTIC) \\
$^3$Marine Research Center of Santa Pola (CIMAR) \\
\\
\texttt{*corresponding author: ismael.bevias@ua.es}
}

\begin{document}
\maketitle
\begin{abstract}

Elasmobranch populations are experiencing significant global declines, and several species are currently classified as threatened. Reliable monitoring and species-level identification are essential to support conservation and spatial planning initiatives such as Important Shark and Ray Areas (ISRAs). However, existing visual datasets are predominantly detection-oriented, underwater-acquired, or limited to coarse-grained categories, restricting their applicability to fine-grained morphological classification.

We present the eLasmobranc Dataset, a curated and publicly available image collection from seven ecologically relevant elasmobranch species inhabiting the eastern Spanish Mediterranean coast, a region where two ISRAs have been identified. Images were obtained through dedicated data collection, including field campaigns and collaborations with local fish markets and projects, as well as from open-access public sources. The dataset was constructed predominantly from images acquired outside the aquatic environment under standardized protocols to ensure clear visualization of diagnostic morphological traits. It integrates expert-validated species annotations, structured spatial and temporal metadata, and complementary species-level information.

The eLasmobranc Dataset is specifically designed to support supervised species-level classification, population studies, and the development of artificial intelligence systems for biodiversity monitoring. By combining morphological clarity, taxonomic reliability, and public accessibility, the dataset addresses a critical gap in fine-grained elasmobranch identification and promotes reproducible research in conservation-oriented computer vision. The dataset is publicly available at https://zenodo.org/records/18549737.

\end{abstract}

\keywords{Elasmobranchs \and Sharks and rays \and Fine-grained image classification \and Biodiversity monitoring \and Marine conservation \and Species recognition \and Computer vision dataset \and Mediterranean Sea \and ISRA \and Marine management tools}

\section{Background \& Summary}

Elasmobranchs constitute a subclass of cartilaginous fishes that includes sharks and batoids (rays, torpedoes and mantas). They inhabit marine waters worldwide and form a group characterized by high morphological diversity and wide size variability.

According to the criteria of the International Union for Conservation of Nature (IUCN) \cite{IUCN}, approximately 1,250 species belong to this subgroup, of which nearly 38\% are currently threatened with extinction. Alarmingly, estimates over time suggest that populations of large sharks have declined by up to 90\%, and although the exact magnitude of these reductions has been questioned, the severity of the observed decline is evident \cite{Gallagher2012}. In particular, in the Mediterranean Sea, population trends within this subgroup show a clear decrease \cite{FERRETTI2008}. This ecosystem is confirmed to host 86 elasmobranch species, and although scientific production in countries across the region has grown exponentially, this increase has not yet translated into effective conservation policies \cite{KOEHLER2025107328}.

Within this context, Important Shark and Ray Areas (ISRAs) emerge as a tool aimed at optimizing spatial planning through the identification of three-dimensional habitat portions that are critical for one or more elasmobranch species and that hold management potential for their conservation \cite{ISRAS}. In this framework, both the recording and monitoring of sightings and the development of tools capable of identifying specific species with high precision are essential, a challenge further aggravated by the high morphological similarity among related species.

\begin{figure}[ht]
    \centering
    \includegraphics[width=0.5\linewidth, trim=0 22em 32em 0, clip]{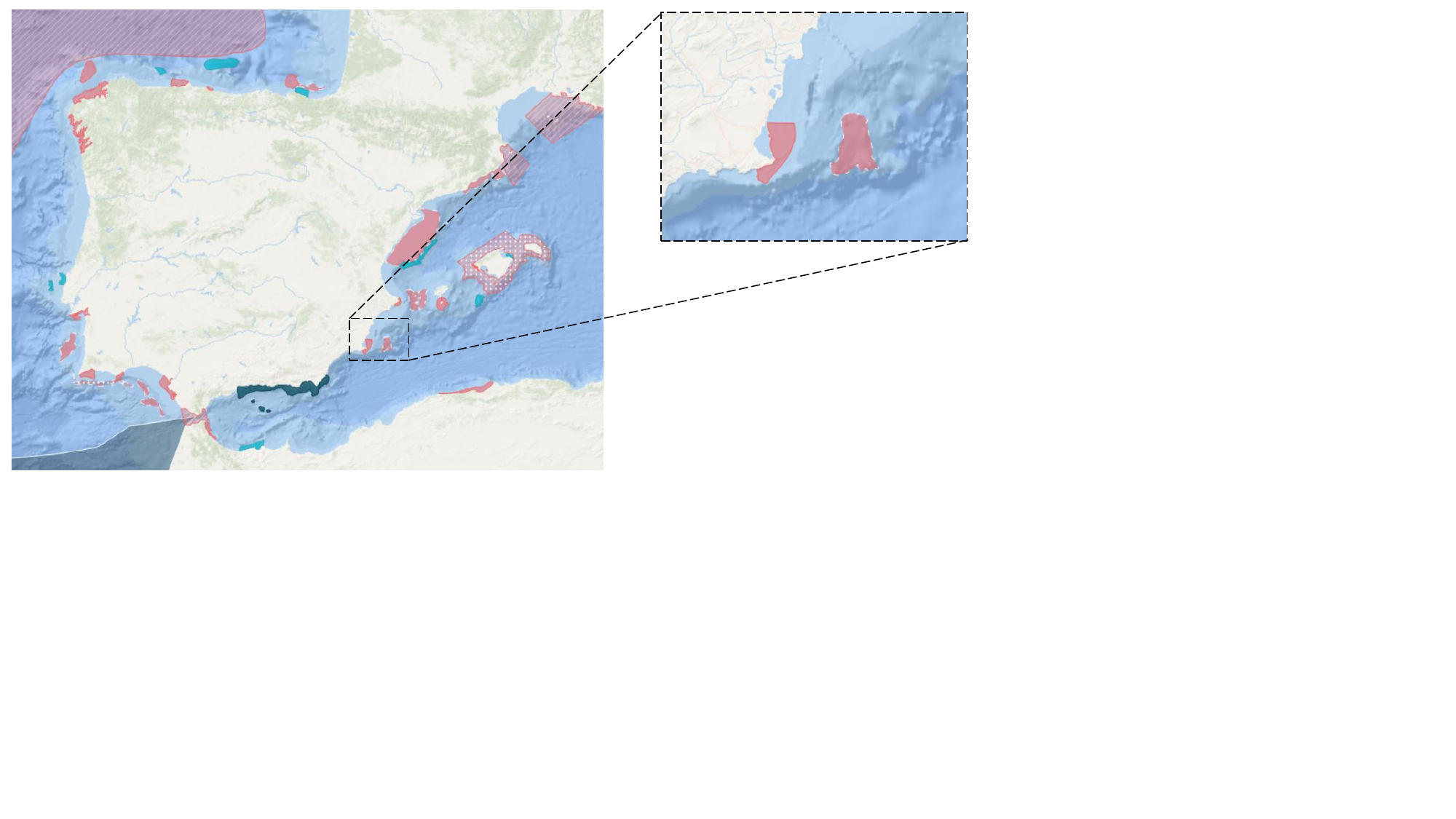}
    \caption{Study area along the eastern Spanish Mediterranean coast, where two ISRAs have been recognized \cite{ISRA_Atlas}.}
    \label{fig:study_area}

\end{figure}

This article presents the \textbf{eLasmobranc Dataset}, a curated image collection focused on the most frequent elasmobranch species along the eastern Spanish Mediterranean coast, a region where multiple Important Shark and Ray Areas (ISRAs) have been designated by the IUCN \cite{pozo2025important,Jabado2023ISRAsMedBlack} (see Figure~\ref{fig:study_area}). Although several ISRAs are recognized within this geographic area, the present study focuses specifically on two of them, where the elasmobranch species included in this dataset were recorded. The dataset comprises seven species (five sharks and two rays), several of which are underrepresented in the literature, and includes species classified within the highest extinction threat categories according to the IUCN Red List; the complete list of species and their conservation status is provided in Table~\ref{tab:iucn_status}. Images were acquired predominantly out of water, except for justified cases, to ensure clear visibility of diagnostic morphological traits and to minimize distortion or environmental noise. A previous version of this dataset supported the development of a deep learning framework for the identification of elasmobranchs \cite{makeELas}.

\begin{table*}[ht]
\centering
\begin{adjustbox}{width=0.8\textwidth}
\begin{tabular}{llll}
\toprule
Species & Global & Europe & Mediterranean \\
\midrule
\textit{Galeus melastomus} & Least Concern & Least Concern & Least Concern \\
\textit{Scyliorhinus canicula} & Least Concern & Least Concern & Least Concern \\
\textit{Galeorhinus galeus} & Critically Endangered & Critically Endangered & Critically Endangered \\
\textit{Mustelus mustelus} & Endangered & Vulnerable & Vulnerable \\
\textit{Oxynotus centrina} & Endangered & Vulnerable & Critically Endangered \\
\textit{Raja undulata} & Near Threatened & Near Threatened & Near Threatened \\
\textit{Torpedo marmorata} & Vulnerable & Least Concern & Least Concern \\
\bottomrule
\end{tabular}
\end{adjustbox}
\caption{Elasmobranch species included in the eLasmobranc Dataset and their conservation status according to the IUCN Red List (global and regional assessments).}
\label{tab:iucn_status}
\end{table*}

As shown in Table~\ref{tab:shark_datasets}, existing elasmobranch datasets are predominantly focused on shark imagery and oriented toward detection and monitoring tasks rather than fine-grained species-level classification. Most resources rely on observations acquired in uncontrolled underwater environments (e.g., \cite{shark-species-dataset_dataset, mozambique-sharks-rays-identification_dataset, varini2024sharktrackaccurategeneralisablesoftware, whale_shark_id_lila}), where factors such as water turbidity, illumination variability, and occlusions hinder the visibility of diagnostic morphological traits. In contrast, few datasets are explicitly designed for species-level identification under standardized acquisition conditions. Other resources (e.g., \cite{frappi2024elasmobranchs, 10261_326154}) emphasize biodiversity reporting through sightings, taxonomy, geolocation, or biometric records, rather than providing structured visual collections. Moreover, several datasets are restricted to coarse-grained categories (e.g., \cite{lopes2024sharks, bemorekgg_aquarium_fish_segmentation_2023, shark_image_classification_imagescv_2025, ticon_shark_ibmby_dataset_2024, varini2024sharktrackaccurategeneralisablesoftware, whale_shark_id_lila}), while those offering species-level annotations typically focus on common taxa (e.g., \cite{shark-species-dataset_dataset, mozambique-sharks-rays-identification_dataset, larusso94_shark_species_2023}). Only \cite{frappi2024elasmobranchs} pursues a partially aligned objective by addressing underwater species-level identification in the Red Sea; however, its emphasis on video data and the integration of multiple acquisition modalities (aerial surveys, BRUV/dBRUV, ROV, fishing, among others) result in a highly heterogeneous dataset, limiting its applicability to controlled visual classification settings. Finally, access restrictions in some large-scale collections (e.g., \cite{JENRETTE2022101673}) further constrain experimental reproducibility and comparative evaluation.

Overall, the lack of homogeneous, species-centric visual datasets acquired under standardized protocols highlights a clear gap in the available resources for fine-grained elasmobranch identification.

\begin{table*}[ht]
\centering
\begin{adjustbox}{width=\textwidth}
\begin{tabular}{llllll}
\toprule
Dataset & Size & Categories & Annotation & Access & Purpose \\
\midrule

Shark Species* \cite{shark-species-dataset_dataset} & 1546 images & 16 & Bounding Box & Public & Classification \& Detection \\

Mozambique Sharks \& Rays \cite{mozambique-sharks-rays-identification_dataset} & 386 images & 11(9) & Bounding Box & Public & Classification \& Detection \\

Aquarium Tracking \cite{lopes2024sharks} & 53,7 GB (video) & 2 & Bounding Box & Public & Tracking \& Detection \\

Aquarium Fish \cite{bemorekgg_aquarium_fish_segmentation_2023} & 638 images & 7(2) & Masks & Public & Semantic Segmentation \\

Shark Species** \cite{larusso94_shark_species_2023} & 1549 images & 14 & Class Labels & Public & Classification \\

Shark Image Classification \cite{shark_image_classification_imagescv_2025} & 537 images & 1 & Class Labels & Public & Classification \\

Shark Capture Vision \cite{ticon_shark_ibmby_dataset_2024} & 1999 images & 1 & Bounding Box & Public & Detection \\

SharkTrack \cite{varini2024sharktrackaccurategeneralisablesoftware} & 6800 images & 2 & Bounding Box \& Tranking & Public & Tracking \& Detection \\

SharkPulse \cite{JENRETTE2022101673} & 53345 images & 219 & Bounding Box & Private & Classification \& Detection \\

CERSE \cite{frappi2024elasmobranchs} & 2847 obs. (373 files)  & 33 & Taxonomy \& Coordinates & Public & Distribution \& Population \\

MEDLEM \cite{10261_326154} & 3000 registers & 34 & Taxonomy \& Biometrics & Public & Distribution \& Population \\

Whale Shark ID \cite{whale_shark_id_lila} & 7888 images & 1 & Bounding Box & Public & Detection \\

E-Lasmobranc (Proposal) & 1117 images & 7 & Class Labels \& Metadata  & Public & Classification \\

\bottomrule
\end{tabular}
\end{adjustbox}
\caption{Summary of related datasets}
\label{tab:shark_datasets}
\end{table*}

\begin{figure}[ht]
    \centering
    \includegraphics[width=0.8\linewidth, trim=0 22em 0 0, clip]{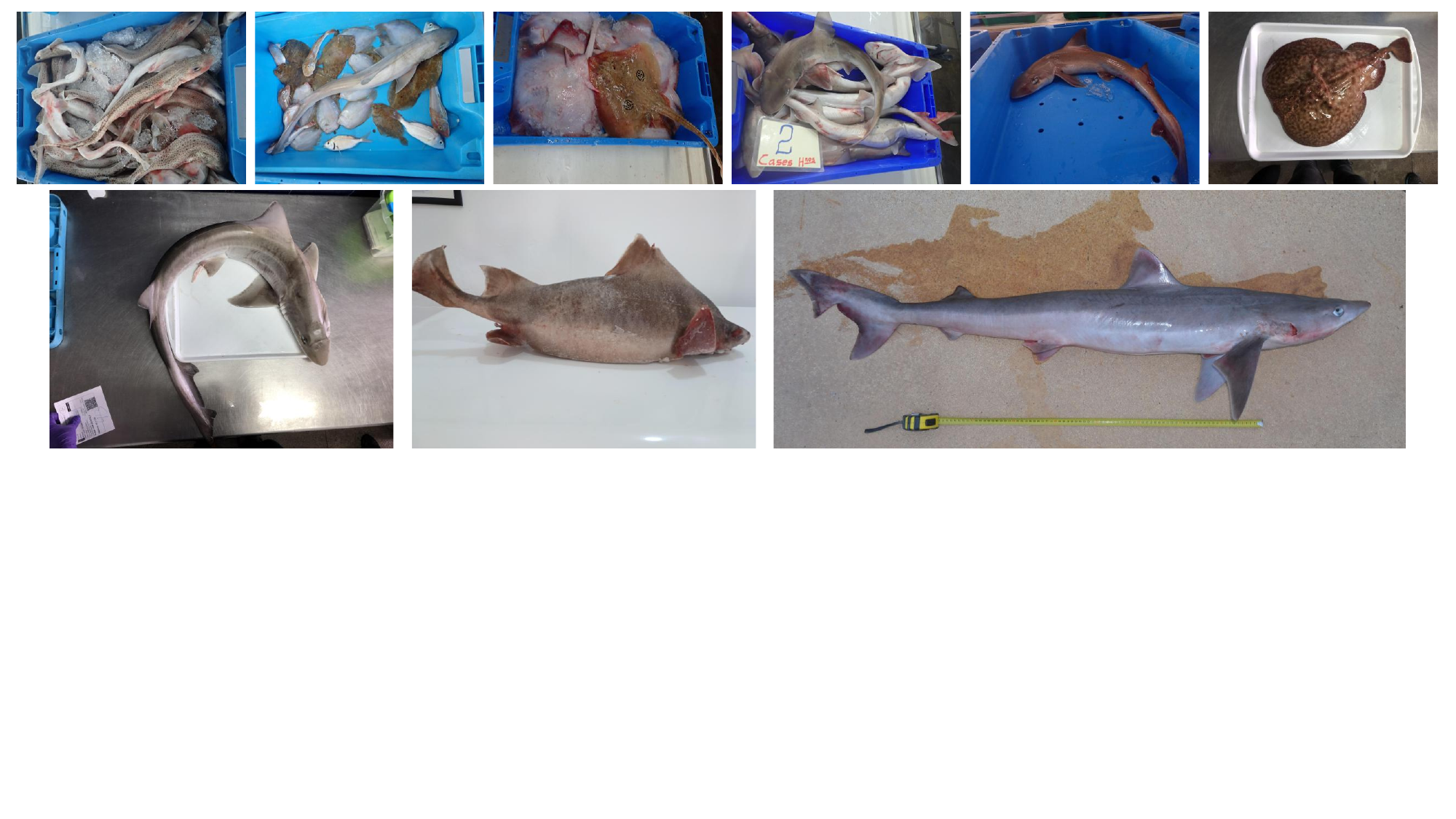}
    \caption{Representative examples from the eLasmobranc Dataset.}
    \label{fig:dataset}

\end{figure}

In contrast, the proposed \textbf{eLasmobranc Dataset} provides a publicly accessible, curated image collection of seven ecologically relevant elasmobranch species from a Mediterranean region. Unlike detection-oriented benchmarks, eLasmobranc is explicitly structured for supervised species-level classification under standardized acquisition conditions, incorporates structured class labels and conservation-relevant metadata, and includes underrepresented and threatened species. By combining regional ecological relevance, morphological clarity, and public availability, the dataset directly supports reproducible fine-grained visual recognition for conservation-oriented computer vision. Representative examples of images from the proposed dataset are shown in Figure~\ref{fig:dataset}.

Each sighting includes at least one image and, in most cases, spatial metadata at the country and administrative subdivision level (community, region, or equivalent), as well as the capture date. Although the dataset focuses on species from the study area, images from different regions worldwide were incorporated to enable statistical analyses and population studies, while also significantly increasing the total volume of available data. Overall, the dataset has been designed to support both marine science research and the development and evaluation of artificial intelligence-based systems.

The images originate both from the research group’s own fieldwork, including targeted capture campaigns and collaborations within the region of interest and nearby areas, and from open-access sources, including large-scale biodiversity and vision datasets (summarized in Table~\ref{tab:global_datasets}) as well as web-based platforms. As seen in works such as \cite{Horn_2018_CVPR} \cite{vendrow2024inquirenaturalworldtexttoimage}, large-scale public datasets are commonly leveraged as source material to construct curated, task-specific collections.  In particular, occurrence records and associated media were retrieved from GBIF \cite{gbif2026globalbiodiversityinformationfacility} and iNaturalist \cite{inaturalist}, which provide extensive biodiversity coverage but required careful filtering to ensure taxonomic consistency and image suitability. As both platforms partially overlap in content, duplicate entries were identified and removed during curation. Additional material was obtained from DeepFish* \cite{garcia-durso2022deepfish}, selectively incorporating samples from the target species that were compatible with the present dataset.

\begin{table*}[ht]
\centering
\begin{adjustbox}{width=\textwidth}
\begin{tabular}{llllll}
\toprule
Dataset & Size & Categories & Annotation & Access & Purpose \\
\midrule

AQUA20 \cite{fuad2025aqua20benchmarkdatasetunderwater} & 8171 images & 20 & Class Labels & Public & Classification \\

FishNet \cite{10377207} & 94532 images & 17357 & Class Labels & Public & Classification \& Detection \\

iNaturalist \cite{inaturalist} & 70M images & 322461 & Class Labels & Public & Classification \\


SCSFish2025 \cite{wang2025scsfish} & 11956 & 30 & Bounding Box & Public & Detection \\

DeepFish* \cite{garcia-durso2022deepfish} & 2602 images & 59 & Bounding Box & Public & Segmentation \\

DeepFish** \cite{saleh2020fish_habitat_dataset} & 40000 & 2 & Bounding Box & Public & Detection \& Segmentation \\

CIFAR-100 \cite{krizhevsky2009learningmultiplelayersfeatures} & 60000 images & 100 & Class Labels & Public & Classification \\

GBIF \cite{gbif2026globalbiodiversityinformationfacility} & 267M records & 5M species & Taxonomy \& Occurrence & Public & Distribution \& Population\\

\bottomrule
\end{tabular}
\end{adjustbox}
\caption{Summary of general datasets}
\label{tab:global_datasets}
\end{table*}

Given the large volume of candidate images available across these sources, an exhaustive case-by-case review was conducted to ensure quality control, correct species labeling, and adequate visibility of diagnostic anatomical traits. While large-scale resources such as AQUA20 \cite{fuad2025aqua20benchmarkdatasetunderwater} and SCSFish \cite{wang2025scsfish} primarily comprise underwater imagery acquired under highly variable environmental conditions, limiting their suitability for fine-grained morphological classification, generic benchmarks such as CIFAR-100 \cite{krizhevsky2009learningmultiplelayersfeatures} contain only very coarse categories (e.g., fish) and provide insufficient anatomical detail. Consequently, only a limited subset of images could be considered appropriate for inclusion. Future extensions will explore systematic integration of large biodiversity repositories such as FishNet \cite{10377207}, with continued emphasis on morphological clarity, taxonomic reliability, and conservation relevance.

\section{Method}

For this study, seven elasmobranch species present along the Spanish Levantine coast were included: \textit{Galeorhinus galeus}, \textit{Galeus melastomus}, \textit{Leucoraja naevus}, \textit{Mustelus mustelus}, \textit{Oxynotus centrina}, \textit{Scyliorhinus canicula} and \textit{Torpedo marmorata}. Images corresponding to each species were obtained predominantly outside the aquatic environment, in order to minimize unwanted visual variability and ensure proper visualization of diagnostic morphological features. 

The image dataset was developed following an exhaustive analysis of data obtained from both external and internal sources.

External sources include large-scale public datasets selected after the review summarized in Table~\ref{tab:global_datasets} (iNaturalist, GBIF and DeepFish*), as well as images collected from web platforms. In all cases, resources were verified to be available under Creative Commons licenses \cite{CC} explicitly allowing reuse (CC0, CC-BY, CC-BY-NC, CC-BY-SA and CC-BY-NC-SA).

\begin{figure}[ht]
\centering
\includegraphics[width=0.3\linewidth]{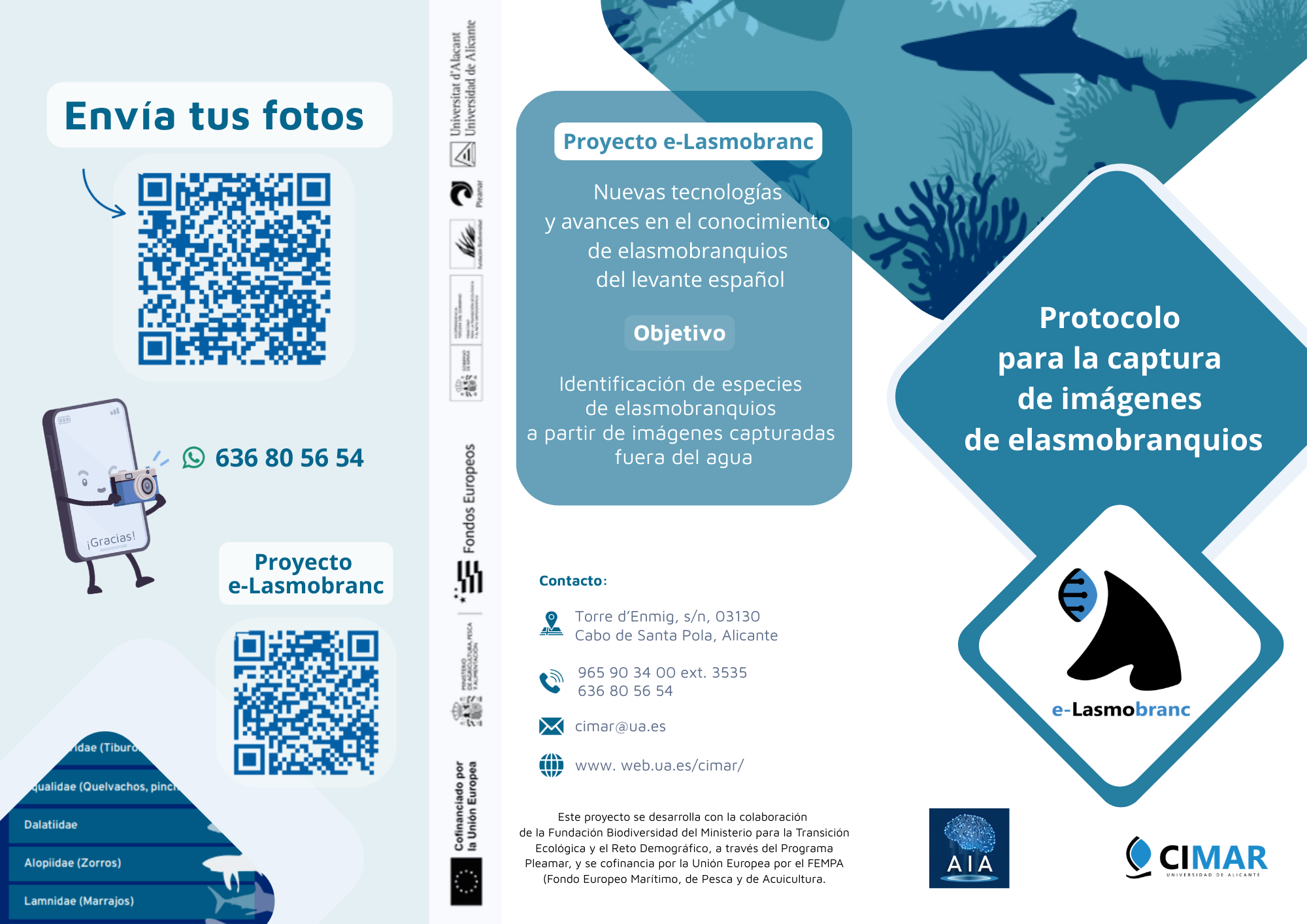}
\hspace{1cm}
\includegraphics[width=0.3\linewidth]{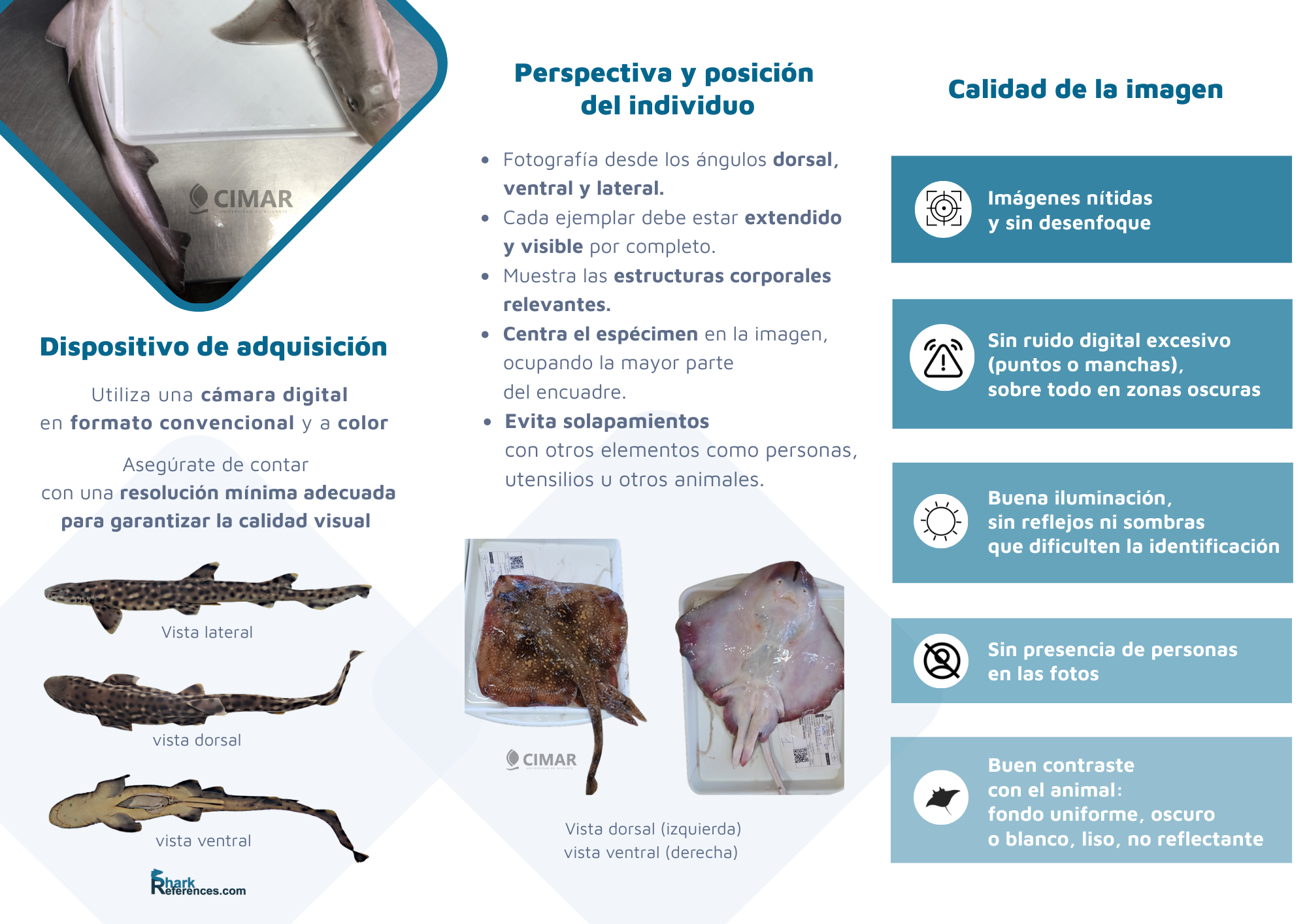}
\caption{Illustrative leaflet developed to disseminate the image acquisition protocol among collaborators. The material is provided in Spanish due to the geographical scope of the study.}
\label{fig:leaflet}

\end{figure}

Internal sources correspond to material generated or directly obtained by the research group, including collaborations with local projects and fish markets. For this purpose, a previously published image acquisition protocol developed by the group was followed \cite{abel2025elasmobranch}, detailing key aspects such as acquisition procedures and criteria related to image quality, perspective and specimen positioning. Additionally, an illustrative leaflet (Figure~\ref{fig:leaflet}) was created to facilitate dissemination of the protocol among collaborators.

The image selection was performed by applying a clearly defined filtering strategy in Table~\ref{tab:curation_criteria}, including both basic criteria and criteria applied by experts.

\begin{table*}[ht]
\centering
\begin{tabular}{p{6cm} p{6cm}}
\hline
\textbf{Basic criteria} & \textbf{Criteria applied by experts} \\
\hline
One of the target elasmobranch species & Duplicates or redundant samples \\
Minimum visual quality threshold & Mislabelled samples \\
Captured outside the aquatic environment & Images lacking minimum diagnostic morphological features   \\
Specimen is fully or sufficiently visible for identification & \\
Contains a single elasmobranch species & \\
Duplicate or redundant images & \\
\hline
\end{tabular}
\caption{Basic curation criteria and criteria applied by experts during dataset construction.}
\label{tab:curation_criteria}

\end{table*}

Finally, the total number of images retained after curation is summarized in the Table~\ref{tab:combined_sources}. In both tables, and more specifically in the massive datasets, the results clearly reflect the extensive filtering effort, which is fundamental to ensure taxonomic reliability and overall data quality.

\begin{table*}[ht]
\centering
\begin{tabular}{lcc}
\hline
\textbf{Source / Dataset} & \textbf{Initial} & \textbf{After expert criteria} \\
\hline
\textbf{Massive datasets (total)} & 11\,519 & 844 \\
\quad iNaturalist & 7\,892 & 766 \\
\quad GBIF & 1\,325 & 74 \\
\quad DeepFish & 2\,602 & 4 \\
\hline
Web sources & -- & 62 \\
Fish market collaborations & 661 & 196 \\
Project collaborations & 240 & 15 \\
\hline
\end{tabular}
\caption{Number of images retained per source and massive dataset after expert validation.}
\label{tab:combined_sources}

\end{table*}

\section{Dataset Composition}

\begin{figure}[ht]
\centering
\includegraphics[width=0.8\linewidth]{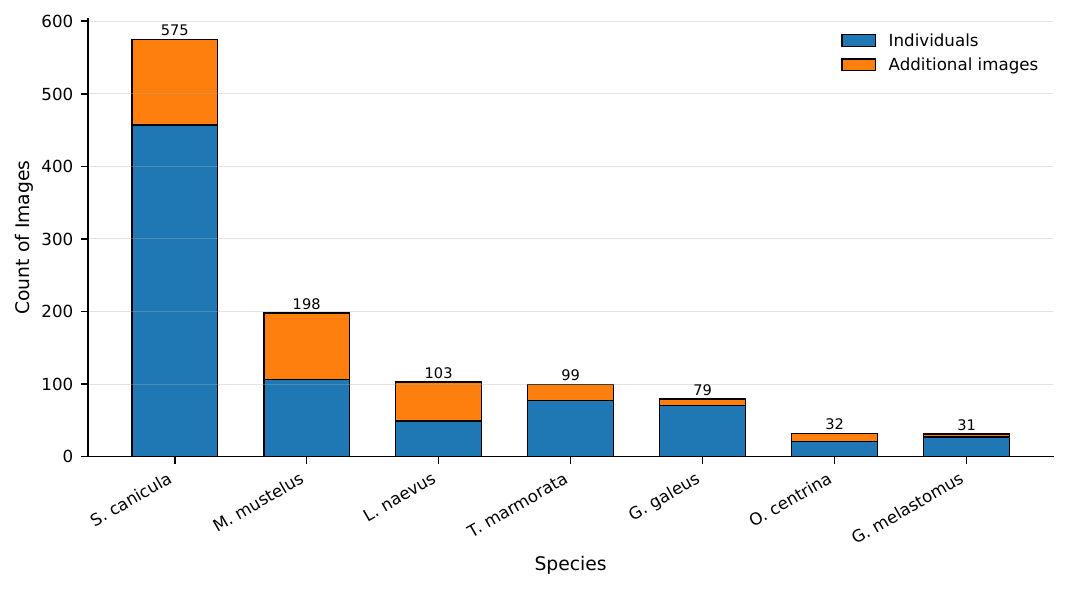}
\caption{Distribution of images and distinct specimens across the elasmobranch species.}
\label{fig:tax}

\end{figure}

\begin{figure}[ht]
\centering
\begin{forest}
for tree={
  grow'=east,
  edge={-latex},
  draw,
  rounded corners,
  align=left,
  font=\small,
  l sep+=12pt,
  s sep+=8pt,
  tier/.option=level
}
[Elasmobranchii
  [Selachii (sharks)
    [Carcharhiniformes
      [Pentanchidae
        [\textit{Galeus melastomus}]
      ]
      [Scyliorhinidae
        [\textit{Scyliorhinus canicula}]
      ]
      [Triakidae
        [\textit{Galeorhinus galeus}]
        [\textit{Mustelus mustelus}]
      ]
    ]
    [Squaliformes
      [Oxynotidae
        [\textit{Oxynotus centrina}]
      ]
    ]
  ]
  [Batoidea (rays)
    [Rajiformes
      [Rajidae
        [\textit{Raja undulata}]
      ]
    ]
    [Torpediniformes
      [Torpedinidae
        [\textit{Torpedo marmorata}]
      ]
    ]
  ]
]
\end{forest}

\caption{Taxonomic hierarchy (class, infraclass, order, family and species) of the elasmobranch species included in the dataset.}
\label{fig:cor}
\end{figure}
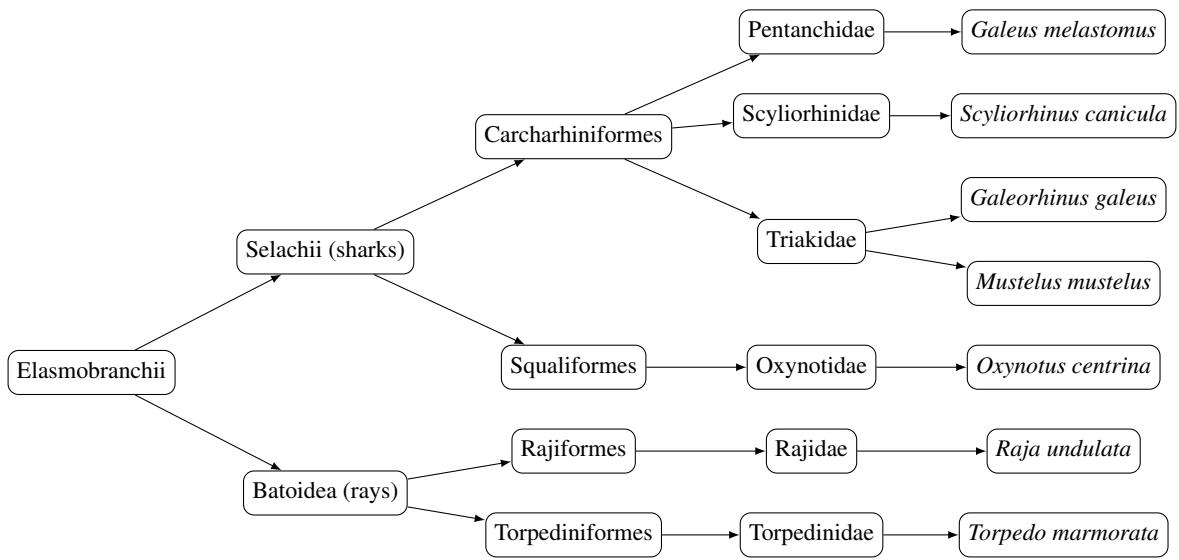

The eLasmobranc Dataset comprises a total of 1\,117 images, of which 902 originate from external sources and 215 from internal sources. It is important to note that the dataset does not exhibit a one-to-one correspondence between images and individuals. Specifically, the 1\,117 images correspond to 807 distinct specimens. The dataset includes the following elasmobranch species: \textit{Galeorhinus galeus}, \textit{Galeus melastomus}, \textit{Leucoraja naevus}, \textit{Mustelus mustelus}, \textit{Oxynotus centrina}, \textit{Scyliorhinus canicula}, and \textit{Torpedo marmorata}, following the distribution shown in Figure~\ref{fig:tax}. Taxonomic relationships among the included species are illustrated in Figure~\ref{fig:cor}, where closer proximity at lower taxonomic levels reflects higher phylogenetic similarity. 

Regarding associated metadata, capture dates were collected with the highest possible temporal resolution (year, month, and day when available), as well as geographic information including country and, when possible, specific area of capture. Out of the total images, 7 lack temporal information at any level, 27 do not include country-level metadata, and 164 images do not provide a more specific geographic area.

\forestset{
  tier/.wrap pgfmath arg={tier #1}{level()}
}

\section{Data Records}

The dataset is publicly available on Zenodo at \url{https://zenodo.org/records/18549737}, and supplementary dataset resources are provided via GitHub at \url{https://github.com/Tech4DLab/eLasmobranc-Dataset}. It consists of a total of 1\,117 images corresponding to the elasmobranch species considered in this study. For all images, spatial and temporal metadata were reviewed or manually annotated. Dates were normalized using the standard DD-MM-YYYY format. Geographic locations were standardized using the Nominatim geocoding service from OpenStreetMap \cite{Nominatim2026}, completing the \textit{country} and \textit{area} fields. The \textit{area} field was assigned by selecting the first available value according to the following priority order: \textit{state}, \textit{archipelago}, \textit{county} and \textit{region}. In cases where it was not possible to associate a capture with a specific country (e.g., open ocean areas), the corresponding marine region or ocean name was used instead.

The dataset is organized into a main directory (ilustred in the Figure~\ref{fig:folderstructure}) that includes: (i) an image folder with independent subdirectories for each species, (ii) an attribution CSV file, (iii) a \textit{citations} text file containing the full references of all data sources, and (iv) an Excel document describing the included species.

\begin{figure}[ht]
\centering

\begin{tcolorbox}[
    width=0.5\linewidth,
    colback=gray!10,
    colframe=gray!50,
    boxrule=0.4pt,
    arc=2mm,
]

\begin{verbatim}
eLasmobranc_Dataset/
|-- data/
|   |-- galeorhinus_galeus/
|   |   |-- GG_001_01.jpg
|   |   |-- ...
|   |   |-- GG_metadata.csv
|   |-- galeus_melastomus/
|   |-- leucoraja_naevus/
|   |-- mustelus_mustelus/
|   |-- oxynotus_centrina/
|   |-- scyliorhinus_canicula/
|   |-- torpedo_marmorata/
|
|-- Elasmobranch taxonomic keys.xlsx
|-- attributions.csv
|-- citations.txt

\end{verbatim}

\end{tcolorbox}

\caption{Folder structure of the eLasmobranc Dataset.}
\label{fig:folderstructure}
\end{figure}

Each image is assigned a unique identifier encoding the species, individual and image index for that individual. For example, the ID \textit{GG\_001\_01} corresponds to the first image of individual 001 of the species \textit{Galeorhinus galeus}. Each species subfolder also contains a dedicated CSV file storing metadata for all images, including identifier, observation date, country and area.

The attribution CSV file stores, for each image, information about the source origin (internal or external), the original license and the corresponding attribution text. The source of origin is also specified and, for external data, the original image identifier is provided. For images obtained from public datasets, the corresponding dataset identifier is included, whereas for images collected from web platforms, the exact source URL is recorded. Internally generated data are distributed under a CC-BY-NC license.

\section{Technical Validation}

All annotations and part of the image acquisition process were performed by a team of marine science experts with extensive experience in elasmobranch research and in the study area. Samples were retained only when the species could be confidently identified and when the distinguishing morphological characteristics were clearly visible. In cases of ambiguity or insufficient visual evidence, the image was discarded following expert-defined criteria.

For externally sourced material, an initial approximate classification was typically available and subsequently verified by the expert team. In contrast, for images acquired directly by the authors, experts were often able to examine specimens in person. In all cases, final labels were assigned through expert consensus, involving multiple specialists to minimize subjectivity and misclassification.
In addition, the distinctive morphological traits of each elasmobranch species were formally documented and incorporated as expert knowledge within informed machine learning methodologies. This expert-derived knowledge has previously been successfully applied in species identification tasks \cite{makeELas}.

\section{Usage Notes}

As the dataset is intended to support multidisciplinary research across marine sciences and computer science, all images are provided in their original format. Since the images originate from heterogeneous sources and were acquired under different settings, a preprocessing step is recommended prior to model training or analysis, depending on the specific task (e.g., resizing or normalization).

In some cases, multiple images correspond to the same individual specimen. These instances are explicitly encoded in the labeling scheme and should be treated jointly. In particular, when splitting the dataset for machine learning experiments, all images belonging to the same individual should be kept within the same subset (e.g., training or testing) to avoid data leakage and to ensure a more realistic and reliable estimation of model performance under different viewpoints or acquisition conditions.

This dataset has been specifically designed to enable population-level studies, spatio-temporal analyses, and cross-regional comparisons, as well as for the training and validation of artificial intelligence systems aimed at improving monitoring, automated identification and decision support in marine conservation contexts.

\section{Competing interests}

The authors declare no competing interests.

\section{Ethics statement}

This study did not involve experimentation on live vertebrates. All images collected correspond to specimens that were already deceased as a result of commercial or recreational fishing activities not conducted for the purposes of this research. No animals were captured, handled, sacrificed, or experimentally manipulated within the framework of this study. Consequently, approval from an animal ethics committee was not required in accordance with national and institutional regulations governing animal experimentation

\section{Acknowledgements}

e-Lasmo2 project is developed with the collaboration of the Biodiversity Foundation of the Ministry for Ecological Transition and the Demographic Challenge, through the Pleamar Programme, and is co-financed by the European Union through the European Maritime, Fisheries and Aquaculture Fund.

\bibliographystyle{unsrt}  
\bibliography{references}

\end{document}